\documentclass[letterpaper]{article} 
\usepackage{aaai23}  
\usepackage{times}  
\usepackage{helvet}  
\usepackage{courier}  
\usepackage[hyphens]{url}  
\usepackage{graphicx} 
\usepackage{natbib}  
\usepackage{caption} 
\usepackage{algorithm}
\usepackage{algorithmic}

\usepackage{framed}
\usepackage{subcaption}
\usepackage{booktabs}
\usepackage{xcolor}
\usepackage{array}
\usepackage{multirow}

\usepackage{newfloat}
\usepackage{listings}
\DeclareCaptionStyle{ruled}{labelfont=normalfont,labelsep=colon,strut=off} 
\floatstyle{ruled}
\newfloat{listing}{tb}{lst}{}
\floatname{listing}{Listing}
\nocopyright

\title{Toward Polar Sea-Ice Classification using Color-based Segmentation and Auto-labeling of Sentinel-2 Imagery to Train an Efficient Deep Learning Model}
\author{
    Jurdana Masuma Iqrah,
    Younghyun Koo,
    Wei Wang,
    Hongjie Xie,
    Sushil Prasad
}
\affiliations{
    The University of Texas at San Antonio\\

    jurdanamasuma.iqrah@utsa.edu, younghyun.koo@utsa.edu, wei.wang@utsa.edu\\hongjie.xie@utsa.edu, sushil.prasad@utsa.edu
}

\usepackage{bibentry}

\begin{document}

\maketitle

\begin{abstract}
Global warming is an urgent issue that is generating catastrophic environmental changes, such as the melting of sea ice and glaciers, particularly in the polar regions. The melting pattern and retreat of polar sea ice cover is an essential indicator of global warming. The Sentinel-2 satellite (S2) captures high-resolution optical imagery over the polar regions.
This research aims at developing a robust and effective system for classifying polar sea ice as thick/snow-covered, young/thin, or open water using S2 images. A key challenge is the lack of labeled S2 training data to serve as the ground truth. We demonstrate a method with high precision to segment and automatically label the S2 images based on suitably determined color thresholds and employ these auto-labeled data to train a U-Net machine model (a fully convolutional neural network), yielding good classification accuracy. 
Evaluation results over S2 data from the polar summer season in the Ross Sea region
of the Antarctic show that the U-Net model trained on auto-labeled data has an accuracy of 90.18\% over the original S2 images, whereas the U-Net model trained on manually labeled data has an accuracy of 91.39\%. Filtering out the thin clouds and shadows from the S2 images further improves U-Net's accuracy, respectively, to 98.97\% for auto-labeled and 98.40\% for manually labeled training datasets. 
\end{abstract}

\section{Introduction}\label{sec:introduction}
In the present age, global warming is a pressing problem that can cause severe environmental alterations, such as sea ice and glacier melting, especially in polar regions. Eventually, it will lead to a rise in sea level, resulting in coastal land loss, a shift in precipitation patterns, increased drought and flood risks, and threats to biodiversity {\cite{serreze2011processes}}. Therefore, assessing the melting and retreat of polar sea ice cover is key as an indicator of global warming.

Numerous satellites are available nowadays that contain large amounts of data about the earth's surface, land elevation, vegetation, etc. In particular, the Sentinel-2 satellite (S2) is an earth observation mission from the Copernicus Program of the European Spatial Agency (ESA) that routinely yields high-resolution optical images over land and coastal seas \cite{drusch2012sentinel}. In addition, Sentinel-2 optical imagery contains images from the polar areas enabling direct analysis of the status of polar sea ice covers.

S2 optical images have a very high resolution of 10m for the R, G, and B bands of the available 13 bands. This resolution is much higher than the Sentinel-1 radar images, which have up to 40m spatial resolution. S2 images are more fine-grained, with more detailed data on the earth's surface, especially on polar sea ice. Sea ice cover classification has been applied on Sentinel-1 radar images using machine learning techniques \cite{park2020classification, boulze2020classification, wang2021arctic}. However, polar sea ice cover classifications on S2 datasets have not been previously carried out. 
On S2 datasets, there is some work on surface classification \cite{campos2020understanding, pelletier2019deep}, sea-ice surface cover classification over S2 data based on manual masking \cite{muchow2021lead} and based on decision trees \cite{wang2021monitoring}. The most closely related work on sea ice monitoring, albeit on a non-polar region, is by \cite{wang2021monitoring}. It is in Liaodong Bay, Bohai sea region, and the sea ice in that non-polar region has characteristics different than the unique polar sea ice. 
However, an early challenge to using S2 images to classify sea ice is the lack of annotated/labeled polar sea ice cover data for training and validation. Because sea ices typically have irregular shapes, and also because of the existence of clouds and shadows in the images, it usually requires careful manual labeling to mark different types of sea ice in the S2 images. This manual labeling is extremely time-consuming and thus not scalable to a larger amount of satellite images. In addition, we observe that the manual labeling by Earth scientists is typically based on the color of the image pixels (e.g., large white areas are usually thick/snow-covered ice). This inspired us to explore color-based segmentation to automatically label S2 polar sea ice images. By observing the collected S2 dataset, we found that the color ranges for polar sea ice and open water are almost constant for the summer season.

This paper reports on the preliminary results of our color-based segmentation/auto-labeling and polar ice classification. We collected about 4000 256x256 pixel images from the polar summer season in the Ross Sea region
of the Antarctic, a relatively small dataset for training machine learning models. Our proposed solution first removes the thin clouds and shadows from the images. Then, we apply our color-segmentation algorithm to automatically label the different sea ice cover images. This step considers color ranges for different ice types and water and produces three different masks for snow-covered/thick ice, thin ice, and open water. Finally, a U-net model is trained on the data automatically annotated by the color-segmentation method to classify polar sea ice and open water or leads.

To validate the effectiveness of auto-labeling and U-net models, we trained two U-net models, one on manually labeled ({\bf U-Net-Man}) and the other on auto-labeled data ({\bf U-Net-Auto}), and validated the models on the same manually labeled validation dataset. The accuracy difference between both models is minimal. 
Our U-Net-Auto model has an accuracy of 90.18\% and 98.97\%, respectively, over the original S2 images and the images with thin clouds and shadows filtered out. In contrast, the U-Net-Man model has an accuracy of 91.39\% and 98.40\%, respectively.\\

The key contributions of this paper are as follows.
\begin{itemize}
    \item A color-based segmentation algorithm to automatically label different sea ice and open water in the polar regions from the S2 images.
    \item A labeled S2 dataset from the polar summer season in the Ross Sea region of the Antarctic with sea ice and water labels for use by the research community. We manually labeled the S2 images of the sea ice cover as well to validate our auto-labeling.
    \item A U-Net model for sea ice cover classification with high accuracy for the polar region using S2 satellite data.
\end{itemize}

Note that although our color-based segmentation method works very well for our current dataset, we do observe that the color ranges employed by the color-based segmentation need to be suitably tailored for S2 images from other polar regions or seasons. 
For future work, we will expand this work-in-progress to more regions and seasons to further validate our hypotheses on the auto-labeling of corresponding training datasets. With that, our expectation is that our U-net model can be trained further to remain robust to these spatial and temporal variations. 

To reproduce our experimental results we made our source code and sample datasets available on GitHub. The link is provided in reference \cite{IqrahSentinel-2Sea-IceClassification2022}.

The remainder of the paper is laid out as follows. Section 2 reviews the key related work. Section 3 describes our proposed methodology for sea ice classification. Section 4 contains the evaluation metrics, experimental results, and a discussion of the proposed methodologies. Finally, in Section 5, we provide concluding remarks and suggest future directions for this ongoing work.

\section{Related Work}\label{sec:rel_works}
Researchers have observed the sea ice melting and retreat, particularly in the arctic, and projected that this amplification is expected to get stronger over time \cite{serreze2011processes}.
Earlier, researchers relied on space-borne satellite data like Sentinel-1 ({\it S1}) Synthetic Aperture Radar (SAR) for extracting sea ice information  \cite{sentinel1sar}. Afterward, Sentinel-2 ({\it S2}) was launched in 2015, offering higher-resolution optical images than S1 SAR images. S2 optical images - that we have employed - have up to 10m high spatial resolution with finer-grain and more detailed sea-ice images compared to S1 with 40m spatial resolution. 
In \cite{park2020classification}, authors proposed Haralick texture features and random forests classifier to retrieve several ice types. However, with around 85\% accuracy, this model is computationally complex and is sensitive to texture noise \cite{8630667} and thermal noise \cite{8126233} in Sentinel-1 data. \cite{boulze2020classification} applied a sea ice types classification using a Convolutional Neural Network (CNN) on Sentinel-1 dual Horizontal-Horizontal (HH) and Horizontal-Vertical (HV) polarization. They successfully trained their CNN and could retrieve four classes of ice: old ice, first-year ice, young ice, and ice-free (open water) with around 90\% accuracy. For ice types' names and codes, they followed the World Meteorological Organization (WMO)  codes \cite{joint2014ice}. They also compared their results with an existing random forest classification \cite{park2020classification} for each ice type and proved theirs to be more efficient based on execution time and less noise sensitivity on SAR data. 
Furthermore, to derive a high-resolution sea ice cover product for the Arctic using S1 dual HH and HV polarization data in extra wide swath (EW) mode, a U-net model was applied in \cite{wang2021arctic}. They used S1 images with 40m spatial resolution to classify sea ice and open water.

\cite{campos2020understanding} used Sentinel-2 time series to classify land use using a 2-BiLSTM recurrent neural network model. 
\cite{muchow2021lead} introduced the application of a sea-ice surface type classification on 20 carefully selected cloud-free Sentinel-2 Level-1C products. To detect {\it leads} (narrow, linear cracks in the ice sheet), they first created five sea-ice surface type classifications (open water, thin ice - nilas, gray sea ice, gray-white sea ice, and sea ice covered with snow); these names are based on the WMO \cite{joint2014ice} for uniformity with other literature. However, they manually masked each five surface classes to get the  Top of Atmosphere (TOA) reflectance value dataset for each mask. 
Another sea-ice monitoring \cite{wang2021monitoring} work on S2 data, in Liaodong Bay, Bohai sea with less than 10\% cloud cover, trained their decision tree on different sea-ice types from Normalized Difference Snow Index (NDSI) and the Normalized Difference Vegetation Index (NDVI). This region has a different type of sea ice (seasonal) than the sea ice in the polar regions.

\section{Methodology}\label{sec:pro_works}
\begin{figure}[ht]
    \begin{framed}
        \centering
        \includegraphics[width=\textwidth]{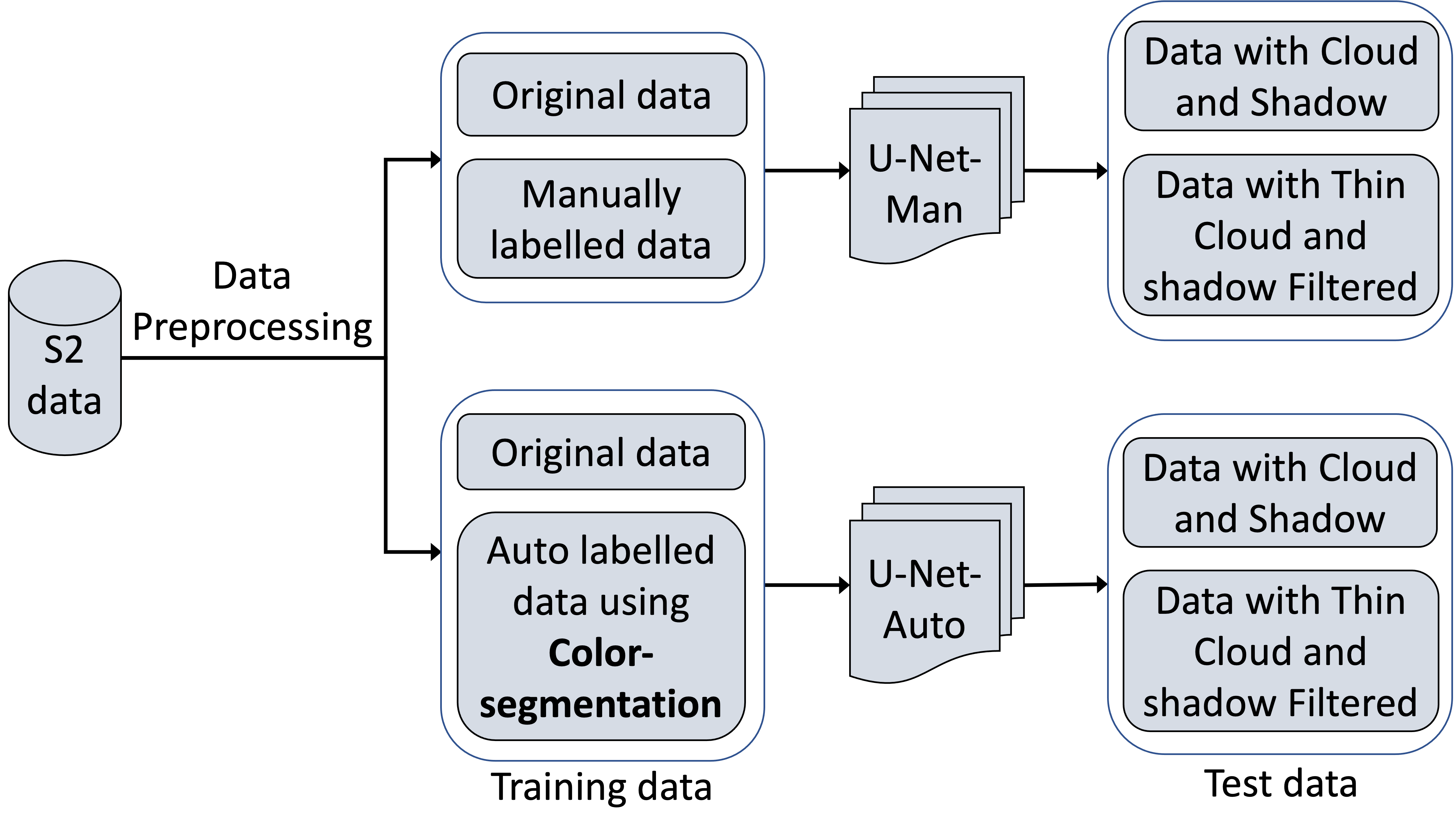}
    \end{framed}
    \caption{Workflow for training and test data preparation and sea ice classification methodology.}
    \label{fig:workflow}
\end{figure}
There are mainly three types of sea ice for classification: thick ice, thin ice, and open water. Identifying these types of sea ice cover will help observe significant changes in the polar sea ice. S2 satellite captures high-resolution optical imagery over land and coastal waters, including the polar sea ice regions. These are more fine-grained images with 10m to 60m spatial resolution. However, these images often include clouds and shadows of clouds that affect the clarity of the corresponding pixels of the image. As a result, the clouds/shadows affect the sea ice segmentation and classification. Therefore, to address this problem, we apply image transformation techniques to remove shadows and thin clouds from the cloudy and shadowy images and store them as thin {cloud and shadow filtered images}. 


A key problem is that there is no labeled sea ice cover data available for training a model. To address this, we employ the workflow as shown in Figure \ref{fig:workflow}. After collecting the S2 imagery for the Ross sea region in the Antarctic, we first manually label the data, mainly for validation purposes. 
Then we also apply color-based segmentation to label different sea ice and open water based on their pixels' color threshold limit values, which the yields auto-labeled S2 sea-ice cover data.
\begin{figure}[ht]
    \begin{framed}
        \centering
        \begin{subfigure}[b]{0.45\textwidth}
            \centering
            \includegraphics[width=\textwidth]{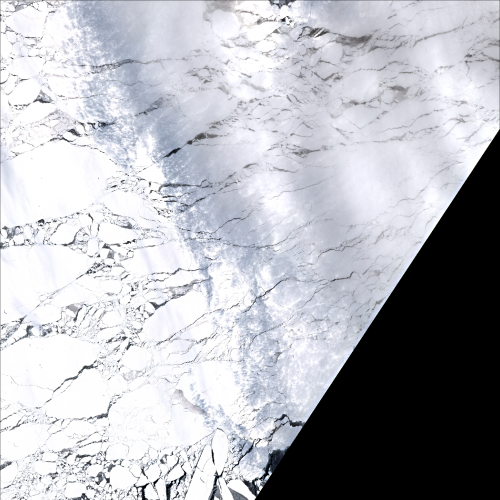}
            \caption{}
            \label{fig:s2_1}
        \end{subfigure}
        \hfill
        \begin{subfigure}[b]{0.45\textwidth}
            \centering
            \includegraphics[width=\textwidth]{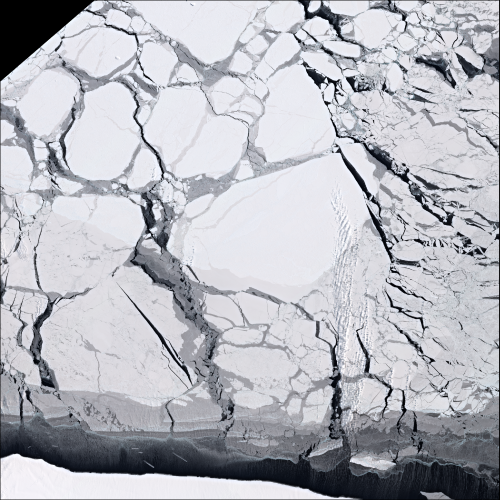}
            \caption{}
            \label{fig:s2_2}
        \end{subfigure}
    \end{framed}
    \caption{Sample Sentinel-2 scenes: (a) with thin cloud/shadow cover, and (b) without cloud or shadows.
    }
    \label{fig:s2 and ross sea}
\end{figure}

\begin{figure}[ht]
    \begin{framed}
        \centering
        \begin{subfigure}[b]{0.3\textwidth}
            \centering
            \frame{\includegraphics[width=\textwidth]{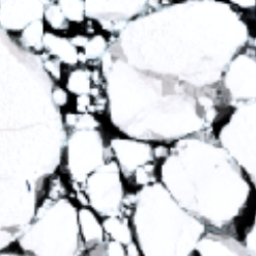}}
            \caption{}
            \label{fig:ori1}
        \end{subfigure}
         \hfill
        \begin{subfigure}[b]{0.3\textwidth}
            \centering
            \frame{\includegraphics[width=\textwidth]{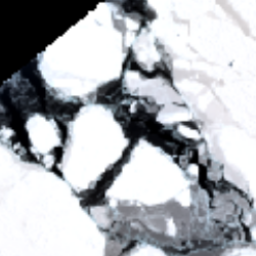}}
            \caption{}
            \label{fig:ori2}
        \end{subfigure}
         \hfill
        \begin{subfigure}[b]{0.3\textwidth}
            \centering
            \frame{\includegraphics[width=\textwidth]{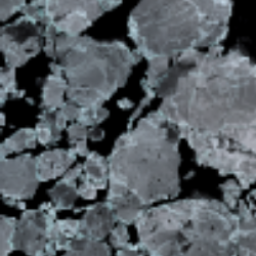}}
            \caption{}
            \label{fig:ori3}
        \end{subfigure}
        \begin{subfigure}[b]{0.3\textwidth}
            \centering
            \frame{\includegraphics[width=\textwidth]{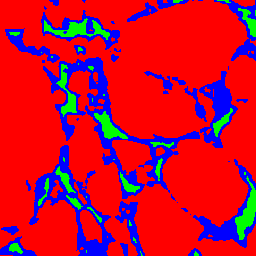}}
            \caption{}
            \label{fig:manu1}
        \end{subfigure}
        \hfill
        \begin{subfigure}[b]{0.3\textwidth}
            \centering
            \frame{\includegraphics[width=\textwidth]{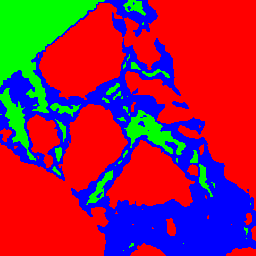}}
            \caption{}
            \label{fig:manu2}
        \end{subfigure}
         \hfill
        \begin{subfigure}[b]{0.3\textwidth}
            \centering
            \frame{\includegraphics[width=\textwidth]{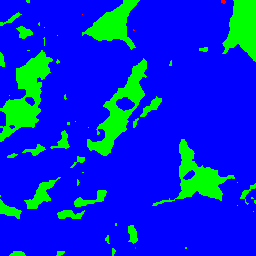}}
            \caption{}
            \label{fig:manu3}
        \end{subfigure}
        \begin{subfigure}[b]{0.9\textwidth}
            \centering
            \includegraphics[width=\textwidth]{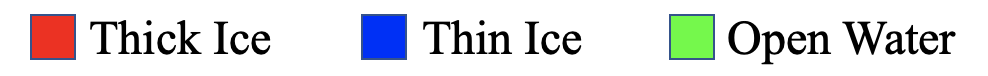}
        \end{subfigure}
        
    \end{framed}
    \caption{Manually-labeled data and their color codes: (a), (b), and (c) are original Sentinel-2 data, (d), (e), and (f) are the respective manually-labeled data. 
    }
    \label{fig:ori and manual}
\end{figure}

Then we train two U-Net models with deep neural network-based architecture, one on manually labeled images (U-Net-Man) and the other on auto-labeled images (U-Net-Auto). Then we compare their accuracy results based on both (i) the original S2 data, including those with cloud and shadows, and (ii) cloud and shadow filtered data to validate our auto-labeling process.  We now describe these in detail.

\subsection{Sentinel-2 Data Collection and Preprocessing}
In our workflow, we first select a spatial and temporal extent.  For our experiments, we choose the well-known spatial region in the Antarctic pole (Ross sea), with spatial extent latitude (south) -70.00 to -78.00 and longitude (west) -140.00 to -180.00. For the temporal extent, we chose November 2019 data, which is the summer season.
Then we collect the S2 satellite imagery for that spatial region and specific time using Google Earth Engine. Each image has 13 available bands; among those available bands, we select bands 2, 3, and 4, representing blue, green, and red, respectively. Each of these bands has a resolution of 10m.

Figure \ref{fig:s2 and ross sea} demonstrates two sample scenes of S2, one with cloud/shadow and one without cloud/shadow. We collected 66 large scenes and split them into over 4,000 images with 256x256 pixel each.
Furthermore, we manually label our Sentinel-2 dataset to test and validate our methodology. We use red for snow-covered/thick ice, blue for thin or young ice, and green for open water regions, as illustrated in Figure \ref{fig:ori and manual}. 

\begin{figure*}[!ht]
        \centering
        \frame{\includegraphics[width=\linewidth]{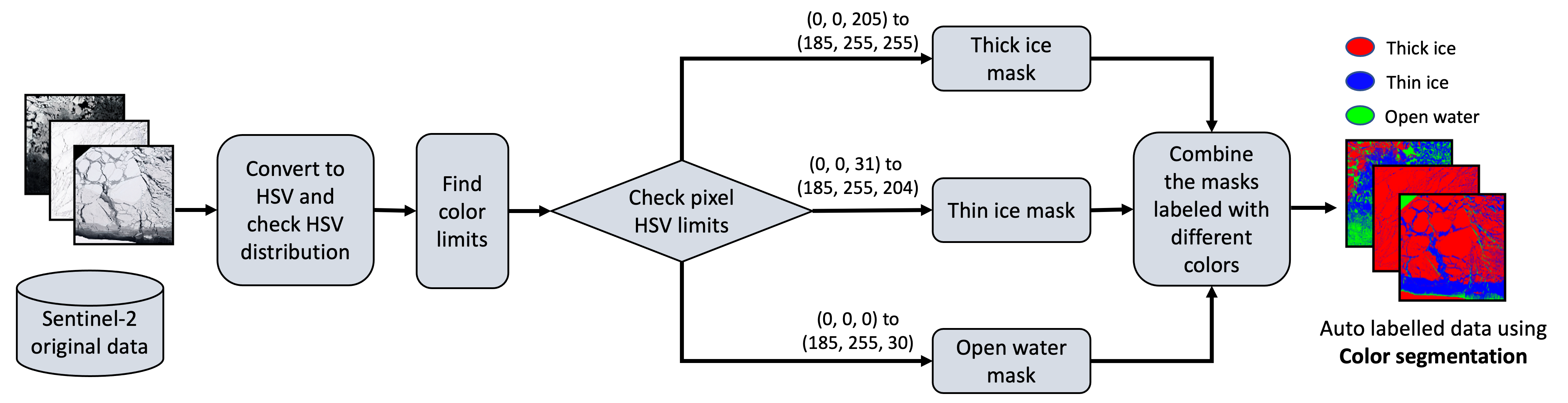}}
    \caption{Color-based segmentation to label thick ice, thin ice, and open water from Sentinel-2 optical imagery.}
    \label{fig:col_seg_archi}
\end{figure*}

\begin{figure}[!htb]
    \begin{framed}
        \centering
        \begin{subfigure}[b]{0.3\textwidth}
            \centering
            \frame{\includegraphics[width=\textwidth]{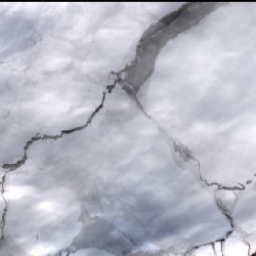}}
            \caption{}
            \label{fig:shdw1}
        \end{subfigure}
        \hfill
        \begin{subfigure}[b]{0.3\textwidth}
            \centering
            \frame{\includegraphics[width=\textwidth]{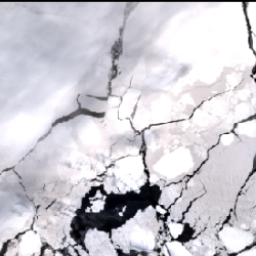}}
            \caption{}
            \label{fig:shdw2}
        \end{subfigure}
        \hfill
        \begin{subfigure}[b]{0.3\textwidth}
            \centering
            \frame{\includegraphics[width=\textwidth]{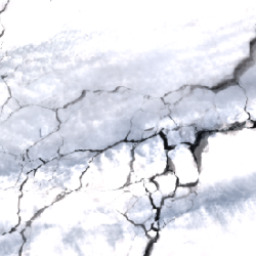}}
            \caption{}
            \label{fig:shdw3}
        \end{subfigure}
        \begin{subfigure}[b]{0.3\textwidth}
            \centering
            \frame{\includegraphics[width=\textwidth]{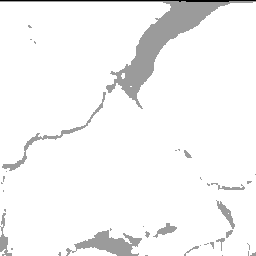}}
            \caption{}
            \label{fig:ref1}
        \end{subfigure}
        \hfill
        \begin{subfigure}[b]{0.3\textwidth}
            \centering
            \frame{\includegraphics[width=\textwidth]{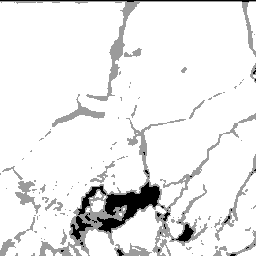}}
            \caption{}
            \label{fig:ref2}
        \end{subfigure}
        \hfill
        \begin{subfigure}[b]{0.3\textwidth}
            \centering
            \frame{\includegraphics[width=\textwidth]{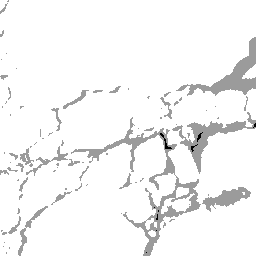}}
            \caption{}
            \label{fig:ref3}
        \end{subfigure}   
    \end{framed}
    \caption{Thin Cloud and Shadow Filtered Dataset: (a), (b) and (c) are Sentinel-2 thin cloudy and shadowy images and (d), (e) and (f) as the corresponding filtered images.}
    \label{fig:shadow and cloud}
\end{figure}

\subsection*{Filtering Out the Thin Clouds and Shadows}
Sentinel-2 images are often significantly affected by thick and thin shadows and cloud covers. As a result, this cloud cover hampers surface observation of the sea ice cover in the polar region. When shadows and clouds are thick, it is difficult to remove those from the image entirely without the corresponding reference shadow, and cloud-free ground truth images of the same region from the same or different satellite image \cite{sarukkai2020cloud, meraner2020cloud}. The problem is that S2 re-visits the same region at five-day intervals, and sea ice cover can change during these five days. 

However, the ground surface is partially visible through thin shadows and clouds. Radiation and brightness factors  can help remove thin clouds from an image \cite{liu2014thin}. Therefore, we decided to apply image transformations to filter our shadowy and cloudy images. For image transformation, we employ a range of known techniques available in OpenCV library \cite{opencv_library}, including RGB to HSV format conversion, noise filtering, bit-wise operations, absolute difference, Otsu thresholding \cite{xu2011characteristic}, Truncated thresholding \cite{guruprasad2020overview} and Binary thresholding, and min-max normalization. 
In Figure \ref{fig:shadow and cloud}, the thin cloudy, and shadowy images with the corresponding filtered images are shown. For both color-segmentation based labeling and U-net model testing, the resulting cloud-and-shadow filtered dataset helps to produce better accuracy for sea ice and open water label and classification. 

\subsection{Color-based Image Segmentation}
Sentinel-2 color segmentation phase segments and labels different sea ice types according to their corresponding pixel color threshold limits as shown in the workflow in Figure \ref{fig:col_seg_archi}. After converting the imagery to HSV format, according to the HSV value distribution of the sea ice segment from the current data set, we manually determine a generic color range (a lower limit and an upper limit) for each sea ice type by trial and error. 
For our Antarctic Ross sea images from the summer season, the HSV lower and upper values for thick ice are (0, 0, {\it 205}) to (185, 255, {\it 255}), for thin ice are (0, 0, {\it 31}) to (185, 255, {\it 204}), and for open water are (0, 0, {\it 0}) to (185, 255, {\it 30}). 
These boundaries do not intersect with each other and can be easily tested against each pixel. With the identified HSV boundaries, we create different masks for each class.
Finally, to address the image labeling/annotation problem, we combine these masks, each labeled with a different color, to perform the color-based segmentation on the S2 sea ice dataset. 
We thus use color segmentation to obtain the auto-labeled S2 sea-ice cover data for the Ross sea region in the Antarctic. 
 
\subsection{U-Net Model for Sea-Ice Classification}
U-Net is a fast training technique that utilizes data augmentation to better use the available annotated labeled data \cite{ronneberger2015u}. A good property of this convolutional network is that it can be trained end-to-end with a small training dataset. U-net is an efficient semantic segmentation model, and applying a multi-class U-Net in our sea ice dataset resulted in semantically segmented and classified sea ice. 

\begin{figure}[ht]
    \begin{framed}
        \centering
        \includegraphics[width=\textwidth]{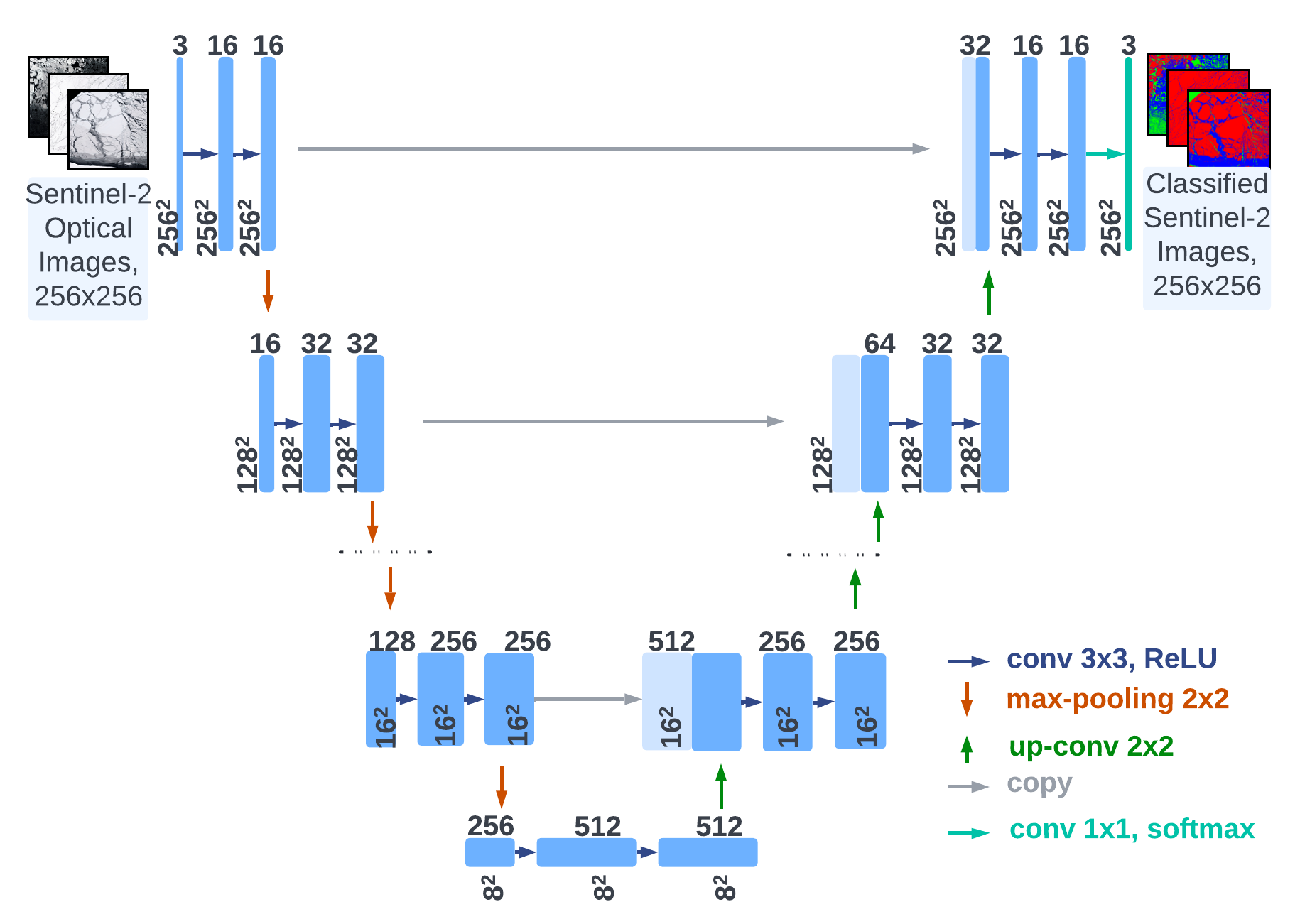}
    \end{framed}
    \caption{U-net model Architecture. 
    }
    \label{fig:unet_archi}
\end{figure}

Our U-Net architecture is shown in Figure \ref{fig:unet_archi}. It consists of three parts: the first one is the contracting path (down-sampling focuses on what the feature is), the second one is the bottleneck, and the third one is the expansion path (up-sampling focuses on where the feature is). Each step of the contraction path consists of two consecutive 3x3 convolutional layers, each with a rectified linear unit (ReLU) \cite{hara2015analysis} followed by a 2x2 max-pooling \cite{nagi2011max} layer with stride 2. The bottleneck step is a single step similar to the down-sampling, except that there is no following max-pooling unit. After that comes the expansion path, where each step consists of (i) first an up-sampling of the feature map,  then (ii) a 2x2 convolution (up-convolution) that halves the number of feature channels, a concatenation with the proportionally cropped feature map from the contracting path, and after that (iii)  two 3x3 convolutions, each followed by a ReLU in the expansive path. Again, due to some loss of boundary pixels in every convolution layer, cropping is necessary. Finally, a 1x1 convolution layer transfers the input features to the desired activation function and the number of classes, which is 3 in our case for thick ice, thin ice, and water.

\subsubsection{U-Net Training:}
For our U-Net model, the input image size is 256x256 pixels. Our model has a total of 28 convolutional layers, including five downsampling steps, one bottleneck step, and five upsampling steps. We use the Adam optimizer  for training and the categorical cross-entropy loss as a loss function for our multi-class model \cite{kingma2014adam}. Our model has some dropout layers in between our convolutional layers. The effect of dropout layers is to regularize the training to avoid overfitting \cite{srivastava2014dropout}. The Adam optimizer continuously improves the loss function on batches of the sample training image set. The optimization is done in epochs, with one epoch being reached when the entire dataset is provided to the neural network for optimization.

\section{Experiments and Results}\label{sec:expr_res}
\subsection{Experimental Setup and Evaluation Metrics}
In this work, we first collected S2 sea ice optical RGB band data using the Google Earth Engine. We collected 66 large scenes from the Ross sea region in the Antarctic. First, we split the large S2 scenes into 256x256 over 4000 images. Then we derived the ground truth/manually labeled data. 
We separated the images into cloudy-shadowy images and cloud shadow filtered images.
Finally, we filtered thin clouds and shadows using OpenCV library image transformation techniques on the cloudy-shadowy data. For the color-based segmentation, as it is a color-limit-based approach, we used the OpenCV library to process this computation. 

As for the U-Net models, first, we divided the dataset into 80\% training dataset and 20\% test dataset. Then we organized the data into batches for the U-Net models using dataloader. Finally, we have used the Adam optimizer; the batch size of 16, 32, and 64; the dropout of 0.1, 0.2, and 0.3 in different convolutional layers; and epochs 50, 70, and 100, respectively, to observe the changes. Our U-Net models have a batch size of 16, and the epoch is 50 for the results reported in the next section. 

To validate the results of our algorithm, the following evaluation metrics over the validation dataset are computed:
\begin{itemize}
    \item \textbf{Classification Accuracy:}
    The overall accuracy is the ratio of the correctly predicted samples over all the samples in the validation set. 
    For the two U-Net models(one trained on a manually labeled dataset and another on the auto-labeled dataset), we evaluate the models using a ground truth validation dataset to find the overall classification accuracy of these two models.  
    
    \item \textbf{Confusion Matrix:}
    For our segmentation model evaluation, we also construct a confusion matrix \cite{ting2017confusion}. The number of samples predicted in category A over the number of samples in category B is specified as an element of the matrix in row A and column B. A complete classification would result in a diagonal confusion matrix, with 100\% on the diagonal and 0\% in the rest of the matrix. Each column adds up to a total of 100\%. This helps understand the model accuracy for classifying each sea ice type individually.
    
\end{itemize}

\subsection{Color-Segmentationn Accuracy}
\begin{figure}[ht]
    \begin{framed}
        \centering
        \begin{subfigure}[b]{0.45\textwidth}
            \centering
            \frame{\includegraphics[width=\textwidth]{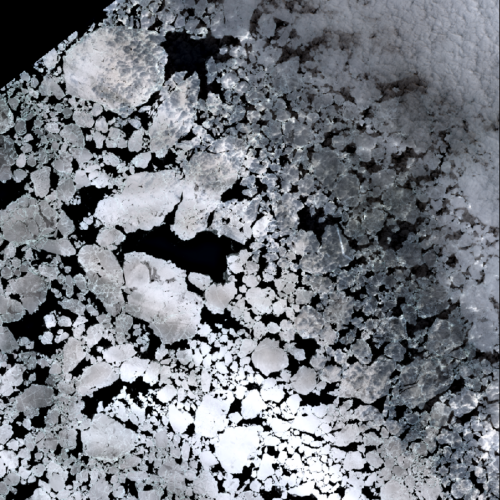}}
            \caption{}
            \label{fig:s2ori}
        \end{subfigure}
        \hfill
        \begin{subfigure}[b]{0.45\textwidth}
            \centering
            \frame{\includegraphics[width=\textwidth]{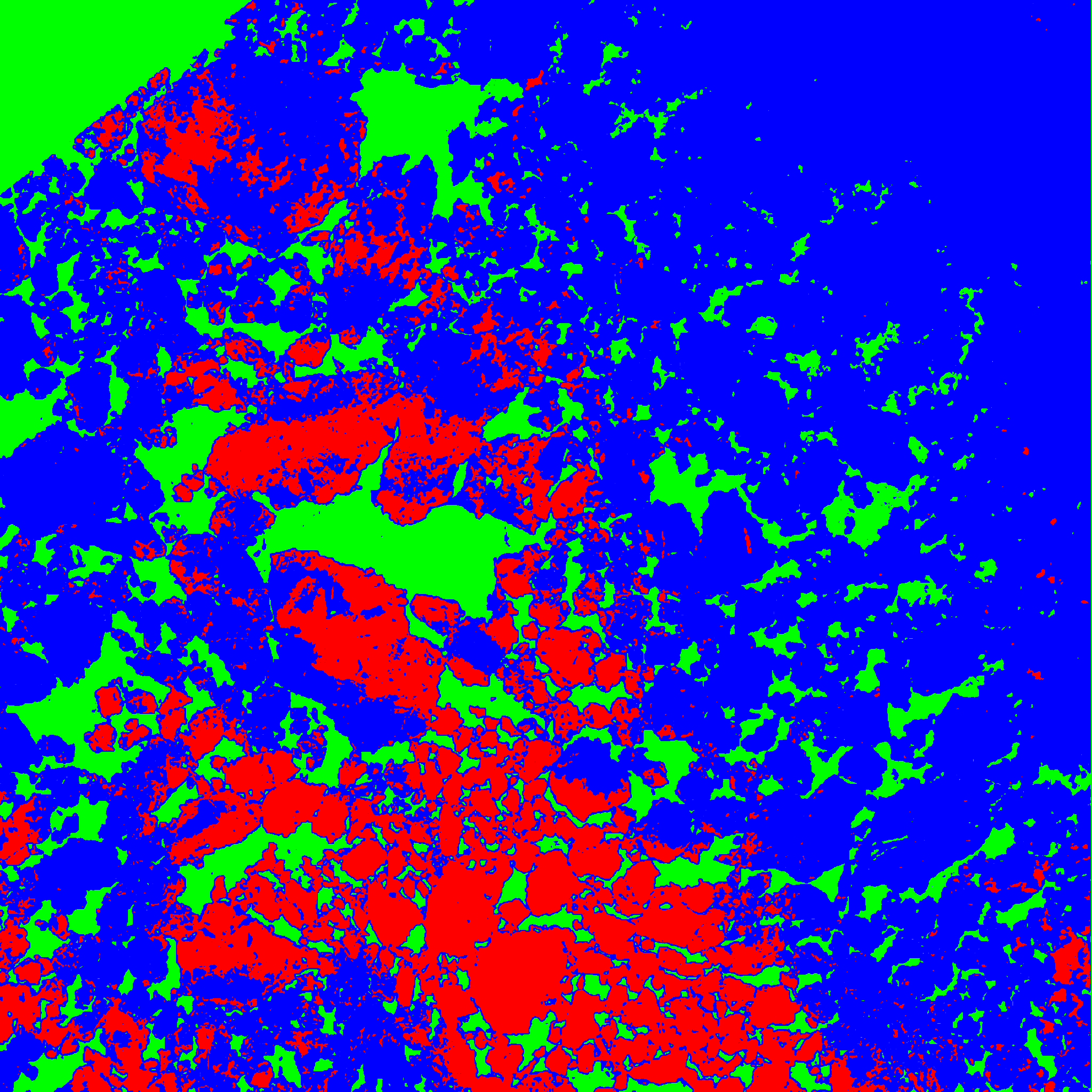}}
            \caption{}
            \label{fig:s2colori}
        \end{subfigure}
        \begin{subfigure}[b]{0.45\textwidth}
            \centering
            \frame{\includegraphics[width=\textwidth]{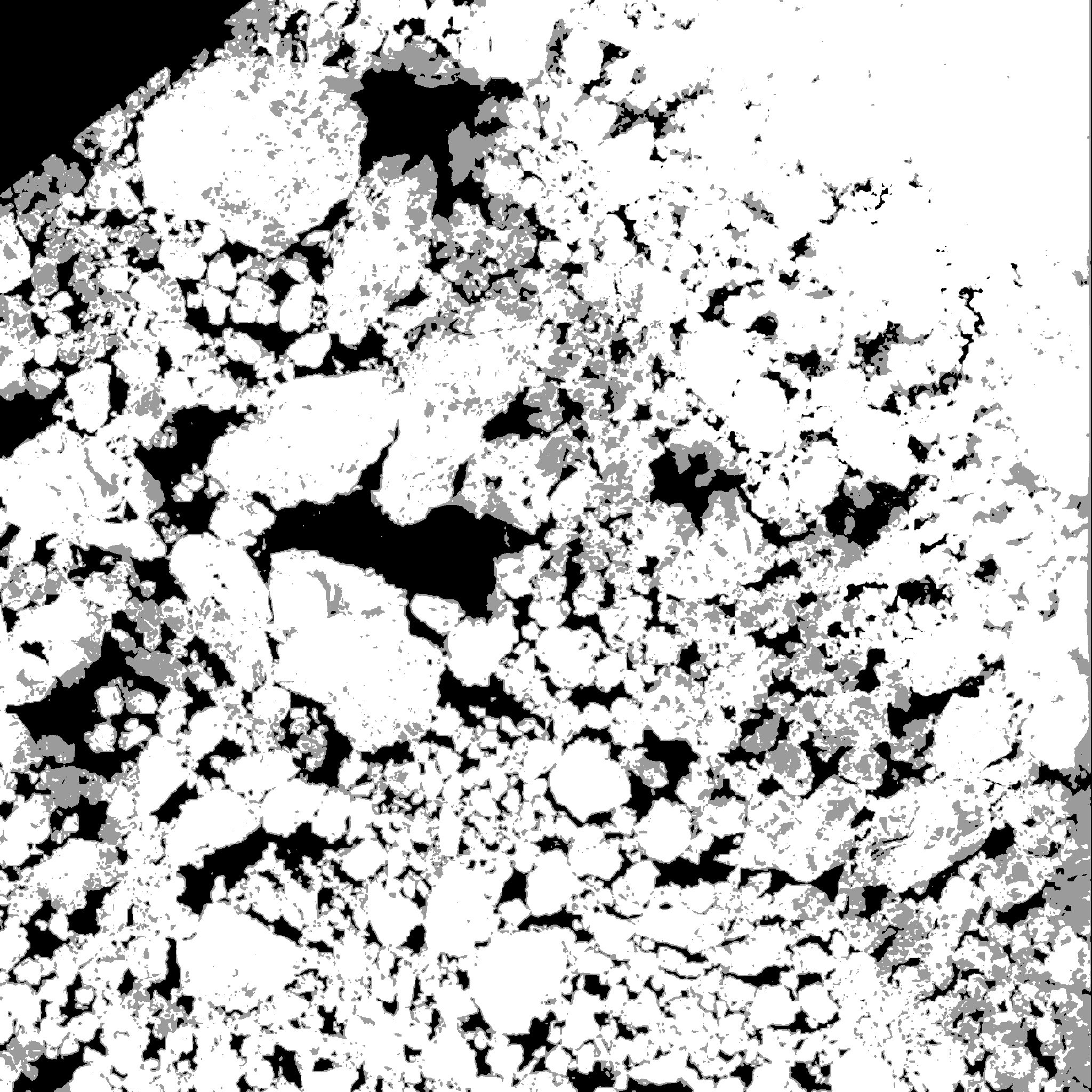}}
            \caption{}
            \label{fig:s2ref}
        \end{subfigure}
        \hfill
        \begin{subfigure}[b]{0.45\textwidth}
            \centering
            \frame{\includegraphics[width=\textwidth]{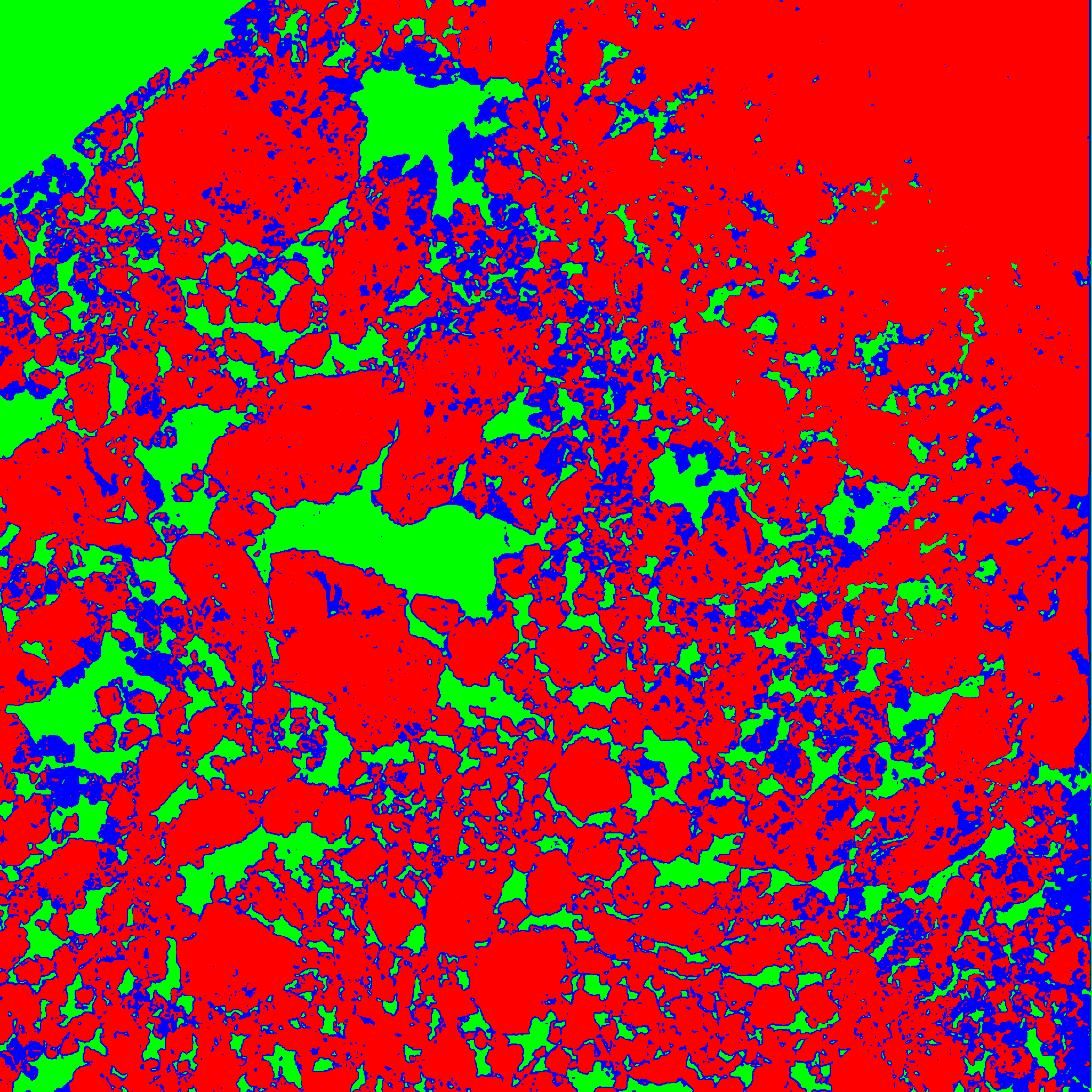}}
            \caption{}
            \label{fig:s2colref}
        \end{subfigure}
        \begin{subfigure}[b]{0.9\textwidth}
            \centering
            \includegraphics[width=\textwidth]{figures/legends.png}
        \end{subfigure}       
    \end{framed}
    \caption{Color-based Segmentation Results: (a) is the Sentinel-2 thin cloudy and shadowy image and (b) is the color-segmented version of (a) with erroneous segmentation in the cloudy/shadowy areas; (c) is (a)'s corresponding thin shadow and cloud filtered image and (d) is the color-segmented version of (c).}
    \label{fig:colsegsample}
\end{figure}

Color-Segmentation is very light in terms of processing time and resource demand. For the original S2 data and the S2 data with clouds and shadows filtered out, we respectively achieved 89\% and 99.64\% Structural Similarity Index (SSIM) \cite{ma2017multi} precision over the manually labeled data. Some sample results of the color segmentation approach are shown 
in Figure \ref{fig:colsegsample}.

We note that the color limits for color-segmentation are not independent of different regions and seasons. For the partial night season of the Antarctic, we had to change the color threshold brightness values manually to label those data to regain accuracy (not reported here). Likewise, the same color limits may not work for different regions of sea ice labeling, and a manual color limit setup may be needed in those cases. On the other hand, we expect our machine model to be robustly trained on the auto-labeled data over various seasons and regions of the poles. 

\subsection{U-Net Model Accuracy}
\begin{table}[htb]
\begin{tabular}{ |>{\centering\arraybackslash}p{0.23\textwidth}||>{\centering\arraybackslash}p{0.08\textwidth}||>{\centering\arraybackslash}p{0.08\textwidth}|>{\centering\arraybackslash}p{0.00\textwidth}}
\hline
\textbf{Dataset}                                              & \textbf{U-Net Man} & \textbf{U-Net Auto} \\ \hline \hline
Original S2 images            & 91.39\%                  & 90.18\%                  \\ \hline
S2 images with thin cloud and shadow filtered              & 98.40\%                  & 98.97\%                   \\ \hline
\end{tabular}
\caption{U-Net models sea ice classification accuracy over Sentinel-2 Antarctic Summer datasets. }
\label{tab:val_acc_comp}
\end{table}

The accuracy comparison of the U-Net-Man and U-Net-Auto is represented in Table \ref{tab:val_acc_comp} for S2 images without thin cloud and shadow filter and S2 images with thin cloud and shadow filter. The U-Net-Man has an accuracy of 91.39\% for original S2 images, and the U-Net-Auto has an accuracy of 90.18\%. After we apply the thin cloud and shadow filter to the S2 images, the accuracy of the U-Net-Man and U-Net-Auto increases to 98.40\% and 98.97\%, respectively.
Thus, the U-Net-Man and U-Net-Auto have very similar accuracy for the classification of sea ice cover. However, after thin cloud and shadow filter on S2 images, the classification accuracy increases for both U-Net-Man and U-Net-Auto by about 7\% and 8\%, respectively.

\begin{table}[hbt]
\begin{tabular}{ |>{\centering\arraybackslash}p{0.3\textwidth}||>{\centering\arraybackslash}p{0.06\textwidth}||>{\centering\arraybackslash}p{0.06\textwidth}|>
{\centering\arraybackslash}p{0.06\textwidth}|>
{\centering\arraybackslash}p{0.00\textwidth}}
\hline
\multicolumn{2}{|p{4cm}|}{\centering \textbf{Dataset}}                                                                        & \textbf{U-Net-Man} & \textbf{U-Net-Auto} \\ \hline
\hline
\multicolumn{1}{|p{3.25cm}|}{\multirow{2}{0.19\textwidth}{More than about 10\% cloud and shadow cover}} & original images & 88.74\%         & 79.91\%           \\ \cline{2-4} 
\multicolumn{1}{|p{1cm}|}{}                                                             & filtered images & 98.91\%         & 99.28\%          \\ \hline
\hline
\multicolumn{1}{|p{3.25cm}|}{\multirow{2}{0.19\textwidth}{Less than about 10\% cloud and shadow cover}} & original images & 92.27\%         & 93.60\%          \\ \cline{2-4} 
\multicolumn{1}{|p{1cm}|}{}                                                             & filtered images & 98.23\%         & 98.87\%          \\ \hline
\end{tabular}
\caption{Sentinel-2 sea ice classification validation accuracy comparison over increasing cloud/shadow coverage.}
\label{tab:val_acc_comp2}
\end{table}
\begin{figure}[!hb]
    \begin{framed}
        \centering
        \begin{subfigure}[b]{0.42\textwidth}
            \centering
            \frame{\includegraphics[width=\textwidth]{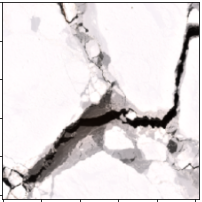}}
            \caption{Original image}
            \label{fig:pred-ori}
        \end{subfigure}
        \hfill
        \begin{subfigure}[b]{0.42\textwidth}
            \centering
            \frame{\includegraphics[width=\textwidth]{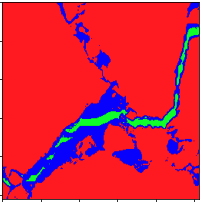}}
            \caption{Ground Truth }
            \label{fig:pred-lbl}
        \end{subfigure}
        \begin{subfigure}[b]{0.42\textwidth}
            \centering
            \frame{\includegraphics[width=\textwidth]{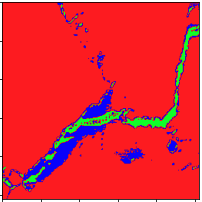}}
            \caption{U-Net-Man Pred}
            \label{fig:pred-man}
        \end{subfigure}
        \hfill
        \begin{subfigure}[b]{0.42\textwidth}
            \centering
            \frame{\includegraphics[width=\textwidth]{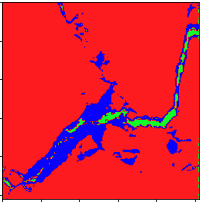}}
            \caption{U-Net-Auto Pred}
            \label{fig:pred-auto}
        \end{subfigure}
        
        
        \begin{subfigure}[b]{0.9\textwidth}
            \centering
            \includegraphics[width=\textwidth]{figures/legends.png}
        \end{subfigure}     
    \end{framed}
    \caption{S2 original image and corresponding manually labeled ground truth image compared to U-Net-Man and U-Net-auto model predictions.}
    \label{fig:validation-auto-man}
\end{figure}
\begin{figure*}[ht]
    \begin{framed}
        \centering
        \includegraphics[width=\textwidth]{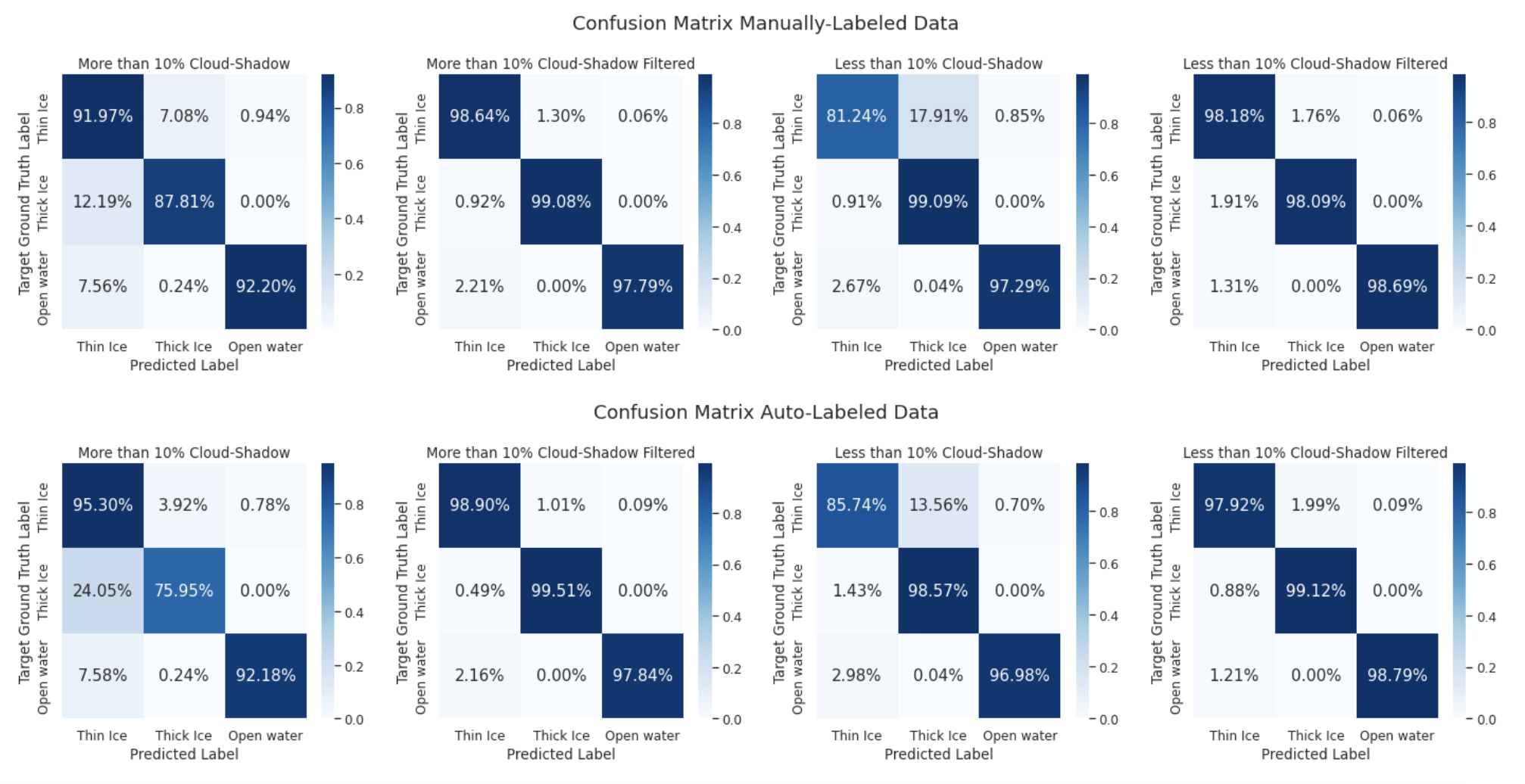}     
    \end{framed}
    \caption{Confusion Matrix of manually labeled and auto-labeled U-net model for thin ice, ice and open water accuracy over cloudy-shadowy, cloud-shadow-removed and cloud-shadow-free.}
    \label{fig:confusion_matrix}
\end{figure*}
Nonetheless, to get a more elaborate comparison, we further divided the S2 validation dataset into (i) more than about 10\% cloud and shadow cover and (ii) less than about 10\% cloud and shadow cover datasets. S2 sea ice classification validation accuracy comparison of U-Net-Man and U-Net-Auto over these two different datasets are represented in Table \ref{tab:val_acc_comp2}. We also included the accuracy of thin cloud and shadow filtered original images along with the original images.
In terms of this cloudy and shadowy dataset, U-Net-Man has better sea ice classification accuracy of 88.74\% compared to U-Net-Auto with 79.91\%. However, after applying our thin cloud and shadow filter, the accuracy of our U-Net-Auto increased to 99.28\%, which is nearly a 20\% increment. On the other hand, U-Net-Man increased to 98.91\% with a 10\% increment.
Again on less than about 10\% cloud and shadow cover original and thin cloud and shadow filtered S2 images, U-Net-Man has 92.27\% and 98.23\% accuracy, respectively, whereas U-Net-Auto has 93.60\% and 98.87\%. For this less cloudy and shadowy dataset, U-Net-Man and U-Net-Auto have similar sea ice classification accuracy, increasing by over 5\% for both models on thin cloud and shadow filtered images. We do not handle thick clouds or shadows here and plan to do that in the future. 

We examine the confusion matrix of thin ice, ice, and open water for the U-Net models with epoch 50 in Figure \ref{fig:confusion_matrix}. It includes the individual accuracy of each class, with thin ice, ice, and open water along the diagonal of the matrix. It also indicates that the thin cloud and shadow filtering improves the individual accuracy of each class along with the overall accuracy shown in Tables \ref{tab:val_acc_comp} and \ref{tab:val_acc_comp2}.
For both U-net models, the accuracy of thick ice, thin ice, and open water is similar (about 98\%) in confusion matrices based on thin cloud and shadow filtered data. 
However, for original data without the thin cloud and shadow filter, when there is more than 10\% cloud and shadow in the image, 12.19\% (U-Net-Man) and 24.05\% (U-Net-Auto) of thick ice is classified as thin ice due to the shadows.
Due to the clouds, the models classify 7.08\% (U-Net-Man) and 3.92\% (U-Net-Auto) thin ice as thick ice; 7.56\% (U-Net-Man) and 7.58\% (U-Net-Auto) open water as thin ice.
In less than 10\% cloudy and shadowy data, thick ice has 99.09\% (U-Net-Man) and 98.57\% (U-Net-Auto) accuracy.  
Due to some presence of thick and thin clouds in the data, thin ice is classified as thick ice by 17.91\% (U-Net-Man) and 13.56\% (U-Net-Auto), and 2.67\% (U-Net-Man) and 2.98\% (U-Net-Auto) open water as thin ice.

Our color-segmented auto-labeled S2 data preparation takes 349.26 seconds for 66 large S2 scenes of 2048x2048 pixels. This time includes our thin cloud and shadow filter method followed by color segmentation to label the S2 images. 
Then for training our U-Net models, we split the original 66 scenes and their corresponding color-segmented thin cloud and shadow-filtered auto-labeled images into 4224 256x256 pixel images. The wall clock numbers of the entire training pipeline for the U-Net-Man model and U-Net-Auto are 1105.88 and 1178.04 seconds, respectively. 
The whole experiment was executed in a Ubuntu 20.4 system with Intel Xeon(R) Silver 4210R 2.40GHz × 20 CPU, 64GB RAM, and NVIDIA Quadro RTX 5000 GPU.

\subsection{Auto-labeling Validation}
Based on the Table \ref{tab:val_acc_comp} and Table \ref{tab:val_acc_comp2}, only slight accuracy difference between the U-Net-Man and U-Net-Auto validates the correctness of our sea ice cover auto-labeling process. 
Figure \ref{fig:validation-auto-man} shows the original S2 image with the manually carried out ground truth label and the prediction generated by the two U-net models, U-Net-Man  and U-Net-Auto.

\section{Conclusions}\label{sec:conclusion}
In this research, we explored the color-based segmentation approach to auto-label the Sentinel-2 satellite images for the purpose of providing annotated data for training deep learning-based sea ice classification models. Our results indicate that color-based segmentation provides quite accurate results in labeling thick ice, thin ice, and open water in the polar regions. Therefore, it could be used to auto-label the S2 satellite polar sea ice cover images in the Antarctic in the summer season. We trained two models -  U-Net-Auto on auto-labeled S2 data and U-Net-Man on manually labeled data - for thick ice, thin ice, and open water classification. Both gave high accuracy, and the accuracy difference between the two models was minimal. 

\paragraph{Future Work:} The color range limits for the color-segmentation are not independent of different regions and seasons. For example, we need different brightness limit control for the partial night season as opposed to the summer season reported here. We intend to extend our color-based segmentation and auto-labeling of corresponding datasets for various seasons and regions.
The U-Net model, on the other hand, is more generic, and we expect to be able to train the model with labeled data over different regions and seasons to achieve a robust sea-ice classification tool independent of seasons and regions.
We also intend to improve our thin cloud and shadow filter to handle thicker clouds and shadows.


\bibliography{ai2ase_2023_iqra.bib}

\end{document}